\documentclass{llncs}
\usepackage{color}
\usepackage{graphicx}
\usepackage{caption}
\usepackage{refstyle}
\usepackage{amsmath}
\usepackage{flexisym}
\usepackage{rotating}
\usepackage{multirow}
\usepackage{hyperref}
\usepackage{rotating}
\usepackage{adjustbox}
\usepackage{blindtext}
\usepackage{nicefrac}
\usepackage{booktabs}
\usepackage{tablefootnote}
\usepackage[table,xcdraw]{xcolor}
\usepackage{tabularx}
\newcommand{\minus}{\scalebox{0.75}[1.0]{$-$}}

\begin{document}
	
	\title{One Single Deep Bidirectional LSTM Network for Word Sense Disambiguation of Text Data}
	\author{Ahmad Pesaranghader\textsuperscript{1,$\dagger$} \and
		Ali Pesaranghader\textsuperscript{2,$\dagger$} \and
		Stan Matwin\textsuperscript{1} \and\\
		Marina Sokolova\textsuperscript{2,3}
	}

	\institute{Institute for Big Data Analytics, Dalhousie University, Halifax, Canada,\and
		School of Computer Science, University of Ottawa, Ottawa, Canada, \and
		School of Epidemiology and Public Health, University of Ottawa, Ottawa, Canada\\\email{\{ahmad.pgh, st837183\}@dal.ca},
		\email{\{apesaran, sokolova\}@uottawa.ca}
	}
	
	\maketitle              

\begin{abstract}
Due to recent technical and scientific advances, we have a wealth of information hidden in unstructured text data such as offline/online narratives, research articles, and clinical reports. To mine these data properly, attributable to their innate ambiguity, a Word Sense Disambiguation (WSD) algorithm can avoid numbers of difficulties in Natural Language Processing (NLP) pipeline. However, considering a large number of ambiguous words in one language or technical domain, we may encounter limiting constraints for proper deployment of existing WSD models. This paper attempts to address the problem of one-classifier-per-one-word WSD algorithms by proposing a single Bidirectional Long Short-Term Memory (BLSTM) network which by considering senses and context sequences works on all ambiguous words collectively. Evaluated on SensEval-3 benchmark, we show the result of our model is comparable with top-performing WSD algorithms. We also discuss how applying additional modifications alleviates the model fault and the need for more training data.

\end{abstract}
\section{Introduction}

Word Sense Disambiguation (WSD) is an important problem in Natural Language Processing (NLP), both in its own right and as a stepping stone to other advanced tasks in the NLP pipeline, applications such as machine translation \cite{vickrey2005word} and question answering \cite{hung2005applying}. WSD specifically deals with identifying the correct sense of a word, among a set of given candidate senses for that word, when it is presented in a brief narrative (surrounding text) which is generally referred to as \textit{context}. Consider the ambiguous word `\textit{cold}'. In the sentence ``\textit{He started to give me a cold shoulder after that experiment}'', the possible senses for cold can be \textit{cold temperature} (S1), \textit{a cold sensation} (S2), \textit{common cold} (S3), or \textit{a negative emotional reaction} (S4). Therefore, the ambiguous word cold is specified along with the sense set \{S1, S2, S3, S4\} and our goal is to identify the correct sense S4 (as the closest meaning) for this specific occurrence of cold after considering - the semantic and the syntactic information of - its context.

In this effort, we develop our supervised WSD model that leverages a Bidirectional Long Short-Term Memory (BLSTM) network. This network works with neural sense vectors (i.e. \textit{sense embeddings}), which are learned during model training, and employs neural word vectors (i.e. \textit{word embeddings}), which are learned through an unsupervised deep learning approach called GloVe (Global Vectors for word representation)\cite{pennington2014glove} for the context words. By evaluating our one-model-fits-all WSD network over the public gold standard dataset of SensEval-3 \cite{mihalcea2004senseval}, we demonstrate that the accuracy of our model in terms of F-measure is comparable with the state-of-the-art WSD algorithms'.

We outline the organization of the rest of the paper as follows. In Section 2, we briefly explore earlier efforts in WSD and discuss recent approaches that incorporate deep neural networks and word embeddings. Our main model that employs BLSTM with the sense and word embeddings is detailed in Section 3. We then present our experiments and results in Section 4 supported by a discussion on how to avoid some drawbacks of the current model in order to achieve higher accuracies and demand less number of training data which is desirable. Finally, in Section 5, we conclude with some future research directions for the construction of sense embeddings as well as applications of such model in other domains such as biomedicine.

\section{Background and Related Work}
Generally, there are three categories of WSD algorithms: supervised, knowledge-based, and unsupervised. Supervised algorithms consist of automatically inducing classification models or rules from labeled examples \cite{zhong2010makes}. Knowledge-based WSD approaches are dependent on manually created lexical resources such as WordNet \cite{miller1995wordnet} and the Unified Medical Language System\footnote{https://www.nlm.nih.gov/research/umls/} (UMLS) \cite{pesaranghader2014word}. Unsupervised algorithms may employ topic modeling-based methods to disambiguate when the senses are known ahead of time \cite{kim2015link}. For a thorough survey of WSD algorithms refer to Navigli \cite{navigli2009word}.

\subsection{Neural Embeddings for WSD}
In the past few years, there has been an increasing interest in training neural word embeddings from large unlabeled corpora using neural networks \cite{collobert2008unified}\cite{mikolov2013efficient}. Word embeddings are typically represented as a dense real-valued low dimensional matrix $\boldsymbol{W}$ (i.e. a \textit{lookup table}) of size $d\times v$, where $d$ is the predefined embedding dimension and $v$ is the vocabulary size. Each column of the matrix is an embedding vector associated with a word in the vocabulary and each row of the matrix represents a latent feature. These vectors can subsequently be used to initialize the input layer of a neural network or some other NLP model. GloVe \cite{pennington2014glove} is one of the existing unsupervised learning algorithms for obtaining these vector representations of the words in which training is performed on aggregated global word-word co-occurrence statistics from a corpus.

Besides word embeddings, recently, computation of sense embeddings has gained the attention of numerous studies as well. For example, Chen et al. \cite{chen2014unified} adapted neural word embeddings to compute different sense embeddings (of the same word) and showed competitive performance on the SemEval-2007 data \cite{navigli2007semeval}.

\subsection{Bidirectional LSTM}

Long Short-Term Memory (LSTM), introduced by Hochreiter and Schmidhuber (1997) \cite{hochreiter1997long}, is a gated recurrent neural network (RNN) architecture that has been designed to address the vanishing and exploding gradient problems of conventional RNNs. Unlike feedforward neural networks, RNNs have cyclic connections making them powerful for modeling sequences. A Bidirectional LSTM is made up of two reversed unidirectional LSTMs \cite{graves2005framewise}. For WSD this means we are able to encode information of both preceding and succeeding words within context of an ambiguous word, which is necessary to correctly classify its sense.

\begin{figure}[hp]
	\begin{adjustbox}{addcode={\begin{minipage}{\width}}{
					\\
					\caption{%
						The single model of deep Bidirectional LSTM for Word Sense Disambiguation of text data.  A series of (left and right) context components are centered around the ambiguous word. The cosine similarities between the context words and the examined sense as the outputs of the first two layers are fed to two LSTM networks with different directions. Then, the concatenated outputs of LSTMs is fed to a (binary) neural network sense classifier consisting of one fully-connected layer and a sigmoid unit. Finally, an argmax over the outputs of all the sigmoids for the set of candidate senses selects the true sense, confirming this sequence of cosine similarities is the best match for the correct sense based on the learned cosine similarities patterns during training of the network.
			}\label{fig:one-model-fits-all}\end{minipage}},rotate=90,center}
		\includegraphics[scale=.26]{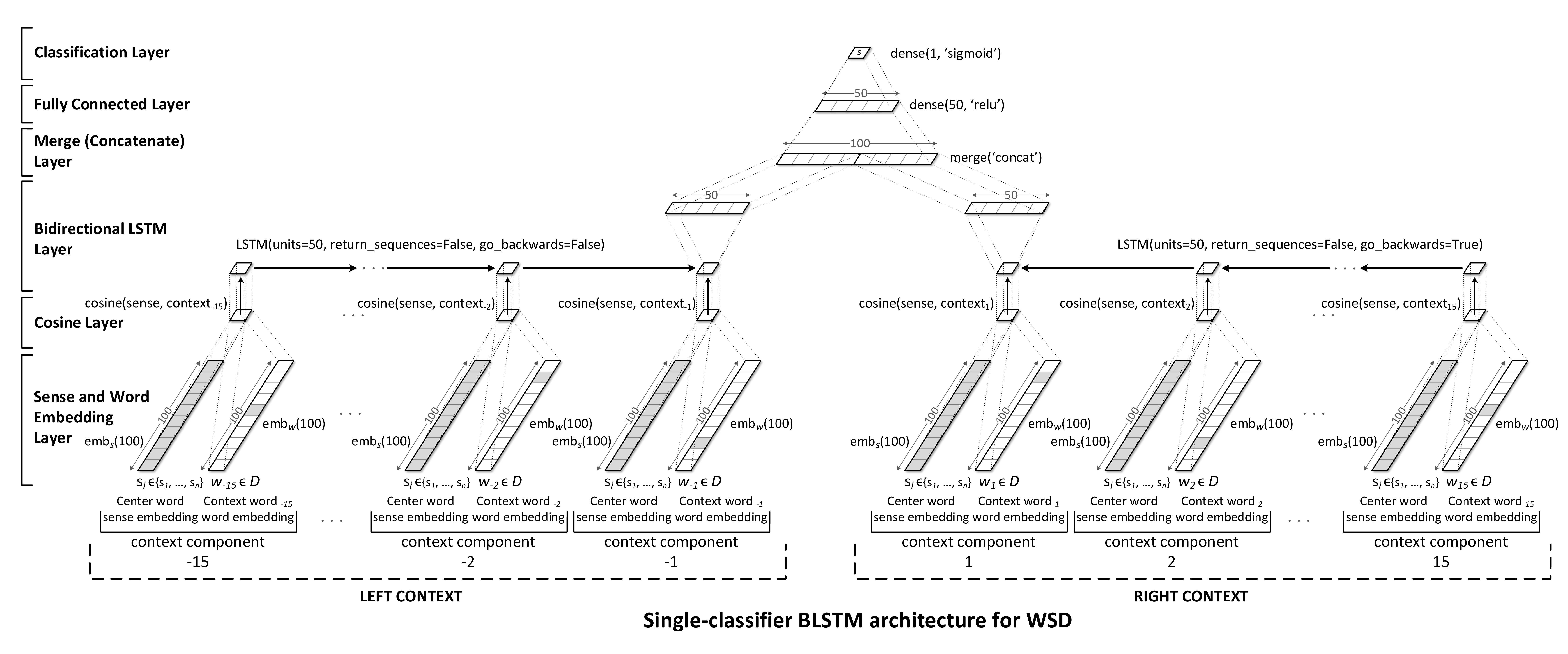}%
		
	\end{adjustbox}
	
\end{figure}

\section{One Single BLSTM network for WSD}
Given a document and the position of a target word, our model computes a probability distribution over possible senses related to that word. The architecture of our model, depicted in Fig. \ref{fig:one-model-fits-all}, consist of 6 layers which are a sigmoid layer (at the top), a fully-connected layer, a concatenation layer, a BLSTM layer, a cosine layer, and a sense and word embeddings layer (on the bottom).

In contrast to other supervised neural WSD networks in which generally a softmax layer - with a cross entropy or hinge loss - is parameterized by the context words and selects the corresponding weight matrix and bias vector for each ambiguous word's senses \cite{kaageback2016word}\cite{taghipour2015semi}, our network shares parameters over all words' senses. While remaining computationally efficient, this structure aims to encode statistical information across different words enabling the network to select the true sense (or even a proper word) in a blank space within a context.

Due to the replacement of their softmax layers with a sigmoid layer in our network, we need to impose a modification in the input of the model. For this purpose, not only the contextual features are going to make the input of the network, but also, the sense for which we are interested to find out whether that given context makes sense or not (no pun intended) would be provided to the network. Next, the context words would be transferred to a sequence of word embeddings while the sense would be represented as a sense embedding (the shaded embeddings in Fig. \ref{fig:one-model-fits-all}). For a set of candidate senses (i.e. $\{s_1, ..., s_n\}$) for an ambiguous term, after computing \textit{cosine similarities} of each sense embedding with the word embeddings of the context words, we expect the sequence result of similarities between the true sense and the surrounding context communicate a pattern-like information that can be encoded through our BLSTM network; for the incorrect senses this premise does not hold. Several WSD studies already incorporated the idea of sense-context cosine similarities in their models \cite{mcinnes2013evaluating}\cite{pedersen2009wordnet}.

\subsection{Model Definition}
For one instance (or one document), the input of the network consists of a sense and a list of context words (left and right) which paired together form a list of context components. For the context \textit{D} which encompasses the ambiguous term $\textit{T}$, that takes the set of predefined candidate senses $\{s_1, ..., s_n\}$, the input for the sense $\textit{s}_i$ for which we are interested in to find out whether the context is a proper match will be determined by Eq. (\ref{eqn:sense-embeddings-input}). Then, this input is copied (next) to $|D|$ positions of the context to form the first pair of the context components.
\begin{align}
\label{eqn:sense-embeddings-input}
\boldsymbol{l}_{i} = \boldsymbol{W}_s^l\cdot\pmb{v}_s(s_i), \; i\in\{1, ..., n\}.
\end{align}
Here, $\pmb{v}_s(s_i)$ is the one-hot representation of the sense corresponding to $s_i \in \{s_1, ..., s_n\}$. A one-hot representation is a vector with dimension $V_s$ consisting of $|V_s| \minus 1$ zeros and a single one which index indicates the sense. The $V_s$ size is equal to the number of all senses in the language (or the domain of interest). Eq. (\ref{eqn:sense-embeddings-input}) will have the effect of picking the column (i.e. sense embeddings) from $\boldsymbol{W}_s^l$ corresponding to that sense. The $\boldsymbol{W}_s^l$ (stored in the sense embeddings lookup table) is initialized randomly since no sense embedding is computed a priori.

Regarding the context words inputs that form the second pairs of context components, at position \textit{m} in the same context \textit{D} the input is determined by:
\begin{align}
\label{eqn:word-embeddings-input}
\boldsymbol{x}_{m} = \boldsymbol{W}_w^x\cdot\pmb{v}_w(w_m), \; m\in\{\nicefrac{-|D|}{2}, ..., -2, -1, 1, 2, ..., \nicefrac{|D|}{2}\}.
\end{align}
Here, $\pmb{v}_w(w_m)$ is the one-hot representation of the word corresponding to $w_m \in D$. Similar to a sense one-hot representation ($V_s$), this one-hot representation is a vector with dimension $V_w$ consisting of $|V_w| \minus 1$ zeros and a single one which index indicates the word in the context. The $V_w$ size is equal to the number of words in the language (or the domain of interest). Eq. (\ref{eqn:word-embeddings-input}) will choose the column (i.e. word embeddings) from $\boldsymbol{W}_w^x$ corresponding to that word. The $\boldsymbol{W}_w^x$ (stored in the word embeddings lookup table) can be initialized using pre-trained word embeddings; in this work, GloVe vectors are used.\\

On the other hand, the output of the network that is examining sense $s_i$ is
\begin{align}
\hat{y}_{s_i} = \sigma( \boldsymbol{W}_{out}\cdot\boldsymbol{h}_{cl}+\boldsymbol{b}_{out}),\: s_i \in \{s_1, ..., s_n\}
\end{align}
where $\boldsymbol{W}_{out}\in R^{1\times 50}$ and $\boldsymbol{b}_{out}\in R$ are the weights and the bias of the classification layer (sigmoid), and $\boldsymbol{h}_{cl}$ is the result of the merge layer (concatenation).

When we train the network, for an instance with the correct sense and the given context as inputs, $\hat{y}_{s_i}$ is set to be \textbf{1.0}, and for incorrect senses they are set to be \textbf{0.0}. During testing, however, among all the senses, the output of the network for a sense that gives the highest value of $\hat{y}_{s_i}$ will be considered as the true sense of the ambiguous term, in other words, the correct sense would be:
\begin{align}
\label{eqn:arg-max}
\arg\max_{s_i}\{\hat{y}_{s_1}, ..., \hat{y}_{s_n}\}, \; s_i \in \{s_1, ..., s_n\}\;.
\end{align}
By applying softmax to the result of estimated classification values, $\{\hat{y}_{s_1}, ..., \hat{y}_{s_n}\}$, we can show them as probabilities; this facilitates interpretation of the results.

Further, the hidden layer ${h}_{cl}$ is computed as
\begin{align}
\label{equation:h}
\boldsymbol{h}_{cl} = ReLU({\boldsymbol{W}_h}\cdot[\boldsymbol{h}_{C_{-1}}^L;\boldsymbol{h}_{C_{+1}}^R]+\boldsymbol{b}_h)
\end{align}
where $ReLU$ means rectified linear unit; $[\boldsymbol{h}_{C_{-1}}^L;\boldsymbol{h}_{C_{+1}}^R]$ is the concatenated outputs of the right and left traversing LSTMs of the BLSTM when the last context components are met. ${\boldsymbol{W}_h}$ and $\boldsymbol{b}_h$ are the weights and bias for the hidden layer.

\subsection{Validation for Selection of Hyper-parameters}
SensEval-3 data \cite{mihalcea2004senseval} on which the network is evaluated, consist of separate training and test samples. In order to find hyper-parameters of the network 5\% of the training samples were used for the validation in advance. Once the hyper-parameters are selected, the whole network is trained on all training samples prior to testing. As to the loss function employed for the network, even though is it common to use \textit{(binary) cross entropy} loss function when the last unit is a sigmoidal classification, we observed that \textit{mean square error} led to better results for the final \textit{argmax} classification (Eq. (\ref{eqn:arg-max})) that we used. Regarding parameter optimization,  RMSprop \cite{hinton2012rmsprop} is employed. Also, all weights including embeddings are updated during training.

\subsection{Dropout and Dropword}
\textit{Dropout} \cite{srivastava2014dropout} is a regularization technique for neural network models where randomly selected neurons are ignored during training. This means that their contribution to the activation of downstream neurons is temporally removed on the forward pass, and any weight updates are not applied to the neuron on the backward pass. The effect is that the network becomes less sensitive to the specific weights of neurons, resulting in better generalization, and a network that is less likely to overfit the training data. In our network, dropout is applied to the embeddings as well as the outputs of the merge and fully-connected layers.

Following the dropout logic, \textit{dropword} \cite{iyyer2015deep} is the word level generalizations of it, but in \textit{word dropout} the word is set to zero while in dropword it is replaced with a specific tag. The tag is subsequently treated just like one word in the vocabulary. The motivation for doing dropword and word dropout is to decrease the dependency on individual words in the training context. Since by replacing word dropout with dropword we observed no change in the results, only word dropout was applied to the sequence of context words during training.  

\section{Experiments}
In SensEval-3 data (\textit{lexical sample task\footnote{http://www.senseval.org/senseval3}}), the sense inventory used for nouns and adjectives
is WordNet 1.7.1 \cite{miller1995wordnet} whereas verbs are annotated with senses from Wordsmyth\footnote{http://www.wordsmyth.net/}. Table \ref{table:SensEval-3-summary} presents the number of words under each part of speech, and the average number of senses for each class.
\begin{table}[]
	\centering
	\caption{Summary of senses in SensEval-3}
	\label{table:SensEval-3-summary}
	\begin{tabular}{@{}llcllc@{}}
		\toprule
		\textbf{Class}      &  & \textbf{Number of words} &  &  & \textbf{Average senses} \\ \midrule
		Nouns      &  & 20              &  &  & 5.8            \\
		Verbs      &  & 32              &  &  & 6.31           \\
		Adjectives &  & 5               &  &  & 10.2           \\ \midrule
		Total      &  & 57              &  &  & 6.47           \\ \bottomrule
	\end{tabular}
\end{table}

As stated, training and test data are supplied as the instances of this task; and the task consist of disambiguating one indicated word within a context.
 
\subsection{Experimental Settings}
The hyper-parameters that were determined during the validation is presented in Table \ref{table:hyper-parameters}. The preprocessing of the data was conducted by lower-casing all the words in the documents and removing numbers. This results in a vocabulary size of $|V|$ = 29044. Words not present in the training set are considered unknown during testing. Also, in order to have fixed-size contexts around the ambiguous words, the padding and truncating are applied to them whenever needed.
\begin{table}[]
	\centering
	\caption{Hyper-parameter used for the experiments and the ranges that were searched during tuning. `-' indicates no tuning was performed on that parameter.}
\label{table:hyper-parameters}
\begin{tabular}{@{}lcc@{}}
	\toprule
	\textbf{Hyper-prameter}          & \textbf{Range searched}        & \textbf{Values used}   \\ \midrule
	Context size                     & {[}10, 100{]} {[}Left, Right{]} & [15 Left, 15 Right]      \\
	Embedding size                   & \{50, 100, 200, 300\}          & 100                    \\
	BLSTM hidden layer size          & {[}50, 300{]}                  & 2*50                   \\
	Dropout on sense/word embeddings  & {[}0, 50\%{]}                  & 20\%                   \\
	Dropout on LSTM outputs          & {[}0, 70\%{]}                  & 50\%                   \\
	Dropout on fully-connected layer & {[}0, 70\%{]}                  & 50\%                   \\
	Word dropout                     & {[}0, 50\%{]}                  & 20\%                   \\
	Sense embedding initialization   & -                              & Random$\:\in\:$unif(-0.1, 0.1) \\
	Word embedding initialization    & -                              & GloVe\tablefootnote{Wikipedia and Gigaword (400K vocab): https://nlp.stanford.edu/projects/glove/} (uncased)        \\ \bottomrule
\end{tabular}
\end{table}

\subsection{Results}
\textit{Between-all-models comparisons - }When SensEval-3 task was launched 47 submissions (supervised and unsupervised algorithms) were received addressing this task. Afterward, some other papers tried to work on this data and reported their results in separate articles as well. We compare the result of our model with the top-performing and low-performing algorithms (supervised). We show our single model sits among the 5 top-performing algorithms, considering that in other algorithms for each ambiguous word one separate classifier is trained (i.e. in the same number of ambiguous words in a language there have to be classifiers; which means 57 classifiers for this specific task). Table \ref{table:between-all-models-comparisons} shows the results of the top-performing and low-performing supervised algorithms.
\begin{table}[]
	\centering
	\caption{F-measure results for SensEval-3 (English lexical samples)}
	\label{table:between-all-models-comparisons}
	\begin{tabular}{@{}lllllc@{}}
		\toprule
		\textbf{Rank} &  & \textbf{Method}                 &  &  & \textbf{F-measure(\%)} \\ \midrule
		1             &  & Multi-classifier BLSTM \cite{kaageback2016word}       &  &  & 73.4               \\
		1             &  & IMS+adapted CW \cite{taghipour2015semi}             &  &  & 73.4               \\
		2             &  & htsa3 \cite{grozea2004finding}                   &  &  & 72.9               \\
		3             &  & IRST-Kernels \cite{strapparava2004pattern}              &  &  & 72.6               \\
		\textbf{4}    &  & \textbf{Our Single-classifier BLSTM} &  &  & \textbf{72.5}      \\
		5             &  & nusels \cite{lee2004supervised}                      &  &  & 72.4               \\ \midrule
		35            &  & IRST-Ties                    &  &  & 58.9               \\
		37            &  & R2D2                         &  &  & 57.2               \\
		39            &  & NRC-Coarse                   &  &  & 48.5               \\
		40            &  & NRC-Coarse2                  &  &  & 48.4               \\
		42            &  & DLSI-UA-LS-SU                &  &  & 44.4               \\ \bottomrule
	\end{tabular}
\end{table}

The first two algorithms represent the state-of-the-art models of supervised WSD when evaluated on SensEval-3. Multi-classifier BLSTM \cite{kaageback2016word} consists of deep neural networks which make use of pre-trained word embeddings. While the lower layers of these networks are shared, upper layers of each network are responsible to individually classify the ambiguous that word the network is associated with. IMS+adapted CW \cite{taghipour2015semi} is another WSD model that considers deep neural networks and also uses pre-trained word embeddings as inputs. In contrast to  Multi-classifier BLSTM, this model relies on features such as POS tags, collocations, and surrounding words to achieve their result. For these two models, softmax constitutes the output layers of all networks. htsa3 \cite{grozea2004finding} was the winner of the SensEval-3 lexical sample. It is a Naive Bayes system applied mainly to raw words, lemmas, and POS tags with correction of the a-priori frequencies. IRST-Kernels \cite{strapparava2004pattern} utilizes kernel methods for pattern abstraction, paradigmatic and syntagmatic information and unsupervised term proximity on British National Corpus (BNC), in SVM classifiers. Likewise, nusels \cite{lee2004supervised} makes use of SVM classifiers with a combination of knowledge sources (part-of-speech of neighboring words, words in context, local collocations, syntactic relations. The second part of the table lists the low-performing supervised algorithms \cite{mihalcea2004senseval}. Considering their ranking scores we see that there are unsupervised methods that outperform these supervised algorithms. 
\\
\\
\textit{Within-our-model comparisons - }Besides several internal experiments to examine the importance of some hyper-parameters to our network, we investigated if the sequential follow of cosine similarities computed between a true sense and its preceding and succeeding context words carries a pattern-like information that can be encoded with BLSTM. Table \ref{table:within-our-model} presents the results of these experiments.
\begin{table}[]
	\centering
	\caption{WSD single-classifier BLSTM with other pieces or hyper-parameters}
	\label{table:within-our-model}
	\begin{tabular}{@{}lllc@{}}
		\toprule
		\textbf{Network (Our Single-classifier)}                 & \textbf{} & \textbf{} & \textbf{F-measure(\%)} \\ \midrule
		Full network in Fig. \ref{fig:one-model-fits-all}                                 &           &           & \textbf{72.5}      \\ \midrule
		BLSTM with reverse directions in Fig. \ref{fig:one-model-fits-all}                            &           &           & 68.9               \\
		BLSTM with a shuffled context                &           &           & 67.3               \\
		Fully-connected layers instead of BLSTM layer            &           &           & 70.2               \\ \midrule
		BLSTM without GloVe for the context (all weights are random)             &           &           & 65.6               \\
		BLSTM without word dropout                               &           &           & 71.1               \\
		BLSTM with a larger context size {[}25 left, 25 right{]} &           &           & 71.4               \\ \bottomrule
	\end{tabular}
\end{table}

The first row shows the best result of the network that we described above (and depicted in Fig. \ref{fig:one-model-fits-all}). Each of the other rows shows one change that we applied to the network to see the behavior of the network in terms of F-measure. In the middle part, we are specifically concerned about the importance of the presence of a BLSTM layer in our network. So, we introduced some fundamental changes in the input or in the structure of the network. Generally, it is expected that the cosine similarities of closer words (in the context) to the true sense be larger than the incorrect senses' \cite{mcinnes2013evaluating}; however, if a series of cosine similarities can be encoded through an LSTM (or BLSTM) network should be experimented. We observe if reverse the sequential follow of information into our Bidirectional LSTM, we shuffle the order of the context words, or even replace our Bidirectional LSTMs with two different fully-connected networks of the same size 50 (the size of the LSTMs outputs), the achieved results were notably less than 72.5\%.

In the third section of the table, we report our changes to the hyper-parameters. Specifically, we see the importance of using GloVe as pre-trained word embeddings, how word dropout improves generalization, and how context size plays an important role in the final classification result (showing one of our experiments).

\subsection{Discussion}
From the results of Table \ref{table:between-all-models-comparisons}, we notice our single WSD network, despite eliminating the problem of having a large number of WSD classifiers, still falls short when is compared with the state-of-the-art WSD algorithms. Based on our intuition and supported by some of our preliminary experiments, this deficiency stems from an important factor in our BLSTM network. Since no sense embedding is made publicly available for use, the sense embeddings are initialized randomly; yet, word embeddings are initialized by pre-trained GloVe vectors in order to benefit from the semantic and syntactic properties of the context words conveyed by these embeddings. That is to say, the separate spaces that the sense embeddings and the (context) word embeddings come from enforces some delay for the alignment of these spaces which in turn demands more training data. Furthermore, this early misalignment does not allow the BLSTM fully take advantage of larger context sizes which can be helpful. Our first attempt to deal with such problem was to pre-train the sense embeddings by some techniques - such as taking the average of the GloVe embeddings of the (informative) definition content words of senses, or taking the average of the GloVe embeddings of the (informative) context words in their training samples - did not give us a better result than our random initialization. Our preliminary experiments though in which we replaced all GloVe embeddings in the network with sense embeddings (using a method proposed by Chen et al. \cite{chen2014unified}), showed considerable improvements in the results of some ambiguous words. That means both senses and context words (while they can be ambiguous by themselves) come from one vector space. In other words, the context would also be represented by the possible senses that its words can take. This idea not only can help to improve the results of the current model, it can also avoid the need for a large amount of training data since senses can be seen in both places, center and context, to be trained.

\section{Conclusion}
In contrast to common one-classifier-per-each-word supervised WSD algorithms, we developed our single network of BLSTM that is able to effectively exploit word orders and achieve comparable results with the best-performing supervised algorithms. This single WSD BLSTM network is language and domain independent and can be applied to resource-poor languages (or domains) as well. As an ongoing project, we also provided a direction which can lead us to the improvement of the results of the current network using pre-trained sense embeddings.

For future work, besides following the discussed direction in order to resolve the inadequacy of the network regarding having two non-overlapping vector spaces of the embeddings, we plan to examine the network on technical domains such as biomedicine as well. In this case, our model will be evaluated on MSH WSD dataset\footnote{https://wsd.nlm.nih.gov/collaboration.shtml} prepared by National Library of Medicine\footnote{https://www.nlm.nih.gov/} (NLM). Also, construction of sense embeddings using (extended) definitions of senses \cite{pesaranghader2015simdef}\cite{pesaranghader2013adapting} can be tested. Moreover, considering that for many senses we have at least one (lexically) unambiguous word representing that sense, we also aim to experiment with unsupervised (pre-)training of our network which benefits form quarry management by which more training data will be automatically collected from the web.

\bibliographystyle{splncs}      
\bibliography{cai}            

\end{document}